\documentclass{ifacconf}

\usepackage{graphicx}      % include this line if your document contains figures
\usepackage{natbib}        % required for bibliography
\usepackage[hyphens]{url}  % DO NOT CHANGE THIS
\usepackage{graphicx} % DO NOT CHANGE THIS
\usepackage{caption} % DO NOT CHANGE THIS AND DO NOT ADD ANY OPTIONS TO IT
\usepackage{algorithm}
\usepackage{algorithmic}
\usepackage{amsfonts, amsmath}
\usepackage{booktabs}   % professional horizontal rules
\usepackage{pifont}     % \ding symbols
\usepackage{xcolor}     % colour the crosses
\usepackage{booktabs}   % nicer horizontal rules (\toprule, \midrule, \bottomrule)
\usepackage{multirow}   % allows \multirow cells
\usepackage{array}      % extended column spec (e.g. >{\centering\arraybackslash}p{..})
\usepackage{graphicx}   % needed for \resizebox (if you scale the table)

\usepackage{siunitx}    % then replace each ‘c’ numeric column by ‘S’ and add
\usepackage{subcaption}  % subfigure support
\usepackage{newfloat}
\usepackage{listings}
\newcommand{\cmark}{\ding{51}}                % ✓
\newcommand{\xmark}{\textcolor{gray}{\ding{55}}} % ✗ in grey
\makeatletter
\newcommand{\DELC}[1]{%
  \@gobble{#1}%
}
\makeatletter
\begin{document}
\begin{frontmatter}

\title{Mitigating Exposure Bias in Risk-Aware Time Series Forecasting with Soft Tokens\thanksref{footnoteinfo}} 

\thanks[footnoteinfo]{Submitted to the IFAC World Congress 2026. This work was partially supported by The LaunchPad for Diabetes program and Capital One.}

% ====== Toggle ======
\newif\ifanon
\anonfalse   % set \anontrue for anonymous version

% ====== Authors ======
\ifanon
    \author[UVA-CS]{Anonymous Authors}
    \address[UVA-CS]{Affiliation withheld for double-blind review.}
\else
    \author[UVA-CS]{Alireza Namazi}
    \author[UVA-CS]{Amirreza Dolatpour Fathkouhi}
    \author[UVA-SDS]{Heman Shakeri}

    \address[UVA-CS]{ 
    Department of Computer Science, University of Virginia, Charlottesville, VA, USA\\
    (e-mails: \{mez4em, aww9gh\}@virginia.edu)}
    \address[UVA-SDS]{ 
    School of Data Science, University of Virginia, Charlottesville, VA, USA\\
    (e-mail: hs9hd@virginia.edu)}
\fi

\begin{abstract}                % Abstract of 50--100 words
Autoregressive forecasting is central to predictive control in diabetes and hemodynamic management, where different operating zones carry different clinical risks. Standard models trained with teacher forcing suffer from exposure bias, yielding unstable multi-step forecasts for closed-loop use. We introduce Soft-Token Trajectory Forecasting (SoTra), which propagates continuous probability distributions (``soft tokens'') to mitigate exposure bias and learn calibrated, uncertainty-aware trajectories. A risk-aware decoding module then minimizes expected clinical harm. In glucose forecasting, SoTra reduces average zone-based risk by 18\%; in blood-pressure forecasting, it lowers effective clinical risk by approximately 15\%. These improvements support its use in safety-critical predictive control.
\end{abstract}


\end{frontmatter}
%===============================================================================

\section{Introduction}

Forecasting plays a central role in safety-critical control applications such as automated insulin delivery and blood-pressure regulation, where the cost of prediction errors varies sharply across operating ranges. In glucose control, for example, a 30\,mg/dL error near the hypoglycemic threshold can trigger severe clinical events, while the same error in the euglycemic range may be benign \citep{kovatchev2000risk}. Similar zone-dependent risk structures arise in hemodynamic management, where deviations in systolic or mean arterial pressure can lead to differing complications depending on the direction of the error and the true value \citep{saugel2018error}. 

Despite this, most forecasting modules used in predictive control and clinical decision support are optimized using uniform losses such as Mean Squared Error (MSE), which treat all deviations equally and ignore the asymmetric consequences of errors \citep{huang2021performance}. Models with strong MSE may still systematically underpredict critical lows or dangerous drops in blood pressure—failures that compromise closed-loop safety even when average accuracy appears high.

\begin{figure}[t]
    \centering
    \includegraphics[width=0.70\linewidth]{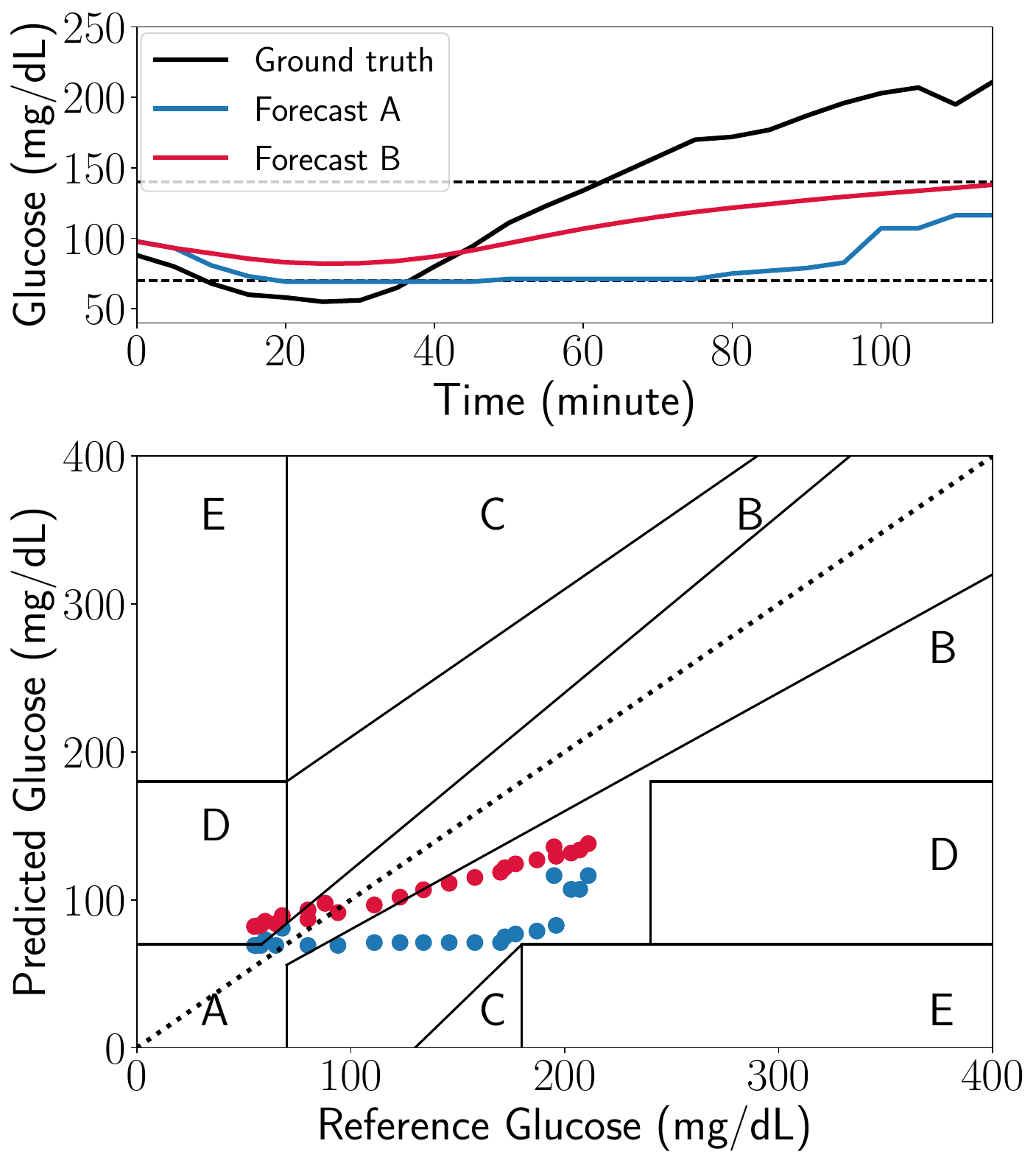}
    \caption{Despite having a lower mean squared error, Forecast~B makes a clinically dangerous error by missing a hypoglycemic event (Zone~D in the Clarke Error Grid). In contrast, Forecast~A stays entirely within Zones~A and B, which indicate clinically safe predictions. The dashed lines indicate the thresholds for hyperglycemia and hypoglycemia. The Clarke Error Grid categorizes blood glucose prediction risk: Zone~A represents accurate estimates, Zone~B benign deviations, while Zones~C--E correspond to increasingly harmful errors. This highlights the importance of risk-aware evaluation in safety-critical forecasting tasks.}
    \label{fig:firstpage}
\end{figure}

A second challenge arises from the widespread use of autoregressive models as forecasting components in control architectures. While autoregressive formulations naturally support multi-step prediction, they are typically trained under \emph{teacher forcing}, conditioning on ground truth rather than their own predictions. This causes \emph{exposure bias} \citep{ranzato2016sequence}: small errors compound over the prediction horizon, yielding unstable trajectories and degrading the reliability required for model-based control. Existing remedies, such as scheduled sampling \citep{bengio2015scheduled}, break differentiability and are difficult to apply to probabilistic forecasters, while search-based decoding (e.g., beam search \citep{collins2004incremental}) is computationally prohibitive in real-time control loops.

We propose Soft-Token Trajectory Forecasting (\textsc{SoTra}), an autoregressive framework designed for risk-sensitive forecasting in clinical control settings. Instead of sampling discrete tokens, \textsc{SoTra} propagates continuous probability vectors (``soft tokens'') through time, preserving uncertainty and enabling end-to-end differentiability of the full prediction trajectory. A domain-specific, risk-aware decoding module then selects predictions that minimize expected clinical harm, aligning the forecasting objective with the utility functions used in closed-loop insulin delivery and blood-pressure control.

Our contributions are:
\begin{itemize}
    \item We introduce \emph{soft-token propagation}, enabling fully differentiable trajectory learning and directly mitigating exposure bias in autoregressive probabilistic forecasting.
    \item We propose a \emph{risk-aware decoding} scheme that decouples model training from utility-driven decision-making, allowing forecasts to be optimized for clinical safety rather than uniform error metrics.
    \item We demonstrate significant reductions in zone-based clinical risk—up to 18\% in glucose prediction and 15\% in blood-pressure forecasting—highlighting SoTra's potential as a forecasting module for safety-critical predictive control systems.
\end{itemize}

\subsection{Related Work}

\subsubsection{\textbf{Time Series Forecasting.}}
Early time series forecasting relied on statistical models \citep{scott2014predicting, hyndman2008forecasting, box2015tsa}, which remain effective for low-dimensional, well-behaved signals. Deep learning models have since emerged as the dominant paradigm \citep{salinas2020deepar, oleg2020nbeats}. Within this class, transformers have become particularly popular due to their scalability and flexibility \citep{Yuqietal-2023-PatchTST}.

For safety-critical domains, uncertainty quantification is essential. One common strategy is to use quantile regression, as in \citet{wen2017multi}, which estimates multiple percentiles. However, each quantile must be trained independently, limiting practicality for complex distributions. Other approaches discretize the output space to allow categorical distribution modeling \citep{ansari2024chronos, gruver2023large}. These classification-style approaches enable modeling of arbitrary distributions, though at the cost of discretization error. An alternative is to generate multiple trajectories through sampling-based generative models \citep{liu2025sundial}, which can capture complex distributions but require multiple forward passes; repeated sampling increases computational cost, limiting real-time use in control loops.

\subsubsection{\textbf{Autoregressive Forecasting.}}
Autoregressive predictors are attractive for multi-step control because they produce coherent trajectories and naturally incorporate uncertainty if framed as a regression-as-classification problem.  
However, these models are usually trained with \emph{teacher forcing}, and suffer from exposure bias \citep{ranzato2016sequence}: they learn to condition on true past values but must condition on self-generated predictions at deployment.  
This mismatch often destabilizes long-horizon forecasts, a critical issue for model-based control.  
Existing remedies—scheduled sampling \citep{bengio2015scheduled}, flipped training \citep{teutsch2022flipped}, or search-based decoding \citep{collins2004incremental}—either break differentiability or add significant inference cost, and do not propagate uncertainty during training.

\subsubsection{\textbf{Zone-Based Risk Assessment.}}
Many clinical variables exhibit zone-dependent risk, where identical numerical errors have different safety implications.  
Error-grid frameworks formalize this idea and are widely used for blood glucose \citep{clarke1987evaluating}, blood pressure \citep{saugel2018error}, and other physiological quantities \citep{hutchings2021quantification, dziorny2023clinical}, where they are used to guide interventions \citep{gardner2006mews, teasdale1974assessment}. Closed-loop systems such as artificial pancreas controllers and automated blood pressure control are increasingly studied in clinical and experimental settings \citep{del2015multicenter, baykuziyev2023closed}. Because model predictive control relies on multi-step predictions to make decisions \citep{el2023intermittent}, incorporating risk-aware forecasts can meaningfully improve its ability to avoid clinically unsafe states.

\section{Soft-Token Trajectory Forecasting}
Safety-critical forecasting domains penalize errors non-uniformly; our goal is therefore to learn calibrated token-level distributions and convert them into point predictions that minimize a risk-weighted cost. Consider a univariate time series with historical observations  
\(\mathcal{X}_h=\{x_1,x_2,\dots,x_T\}\).  
Given a look-back window of length \(T\), we estimate the predictive
distributions for the next $L$ timesteps denoted by  
\(\hat{\mathcal{P}}_f=\{\hat{\boldsymbol{p}}_{T+1},\dots,\hat{\boldsymbol{p}}_{T+L}\}\),
\(\hat{\boldsymbol{p}}_{t}\in\Delta^{V-1}\), from which point forecasts
\(\hat{\mathcal{X}}_f=\{\hat{x}_{T+1},\dots,\hat{x}_{T+L}\}\) are derived.

In many safety-critical tasks, forecast quality is evaluated not by standard statistical losses (e.g., MSE) but by \emph{zone-based risk functions}. In such settings, the $(x_{T+i}, \hat{x})$ plane is partitioned into \(K\) mutually exclusive zones \(\mathcal{Z} = \{Z_1, Z_2, \dots, Z_K\}\), each associated with a risk weight \(w_k\), which quantifies the severity of prediction errors falling within that zone.

Let \(f_r(x, \hat{x}) \in \{w_1, \dots, w_K\}\) denote the zone-based risk function, assigning a penalty to the prediction \(\hat{x}\) given the true value \(x\). The goal of the model is to produce point forecasts that minimize the total expected risk over the forecast horizon:
\begin{equation}
\mathbb{E}[\mathcal{R}] = \sum_{i=T+1}^{T+L} \mathbb{E}_{x_i} \left[ f_r(x_i, \hat{x}_i) \right],
\end{equation}
where \(\mathcal{R}\) reflects the clinical consequences of prediction errors. This motivates \textsc{SoTra}, which integrates zone-aware risk only at the decoding stage. We do not train the model on the risk-aware loss, as this would harm probabilistic calibration and cause compounding bias in an autoregressive setting. Instead, we learn calibrated token distributions and apply risk minimization only when producing point forecasts. \textsc{SoTra} is implemented as a decoder-only autoregressive regression-as-classification model, with values discretized into tokens. It has three components:
\begin{enumerate}
  \item \textbf{Soft-Token Embedding.}
        The soft embedder maps the estimated distribution of the time series at time $t$, $\hat{\boldsymbol{p}}_t$, to the embedding vector $\boldsymbol{e}_t$. This contrasts with conventional token-based approaches that sample a token from the distribution and pass it to the next step. Sampling breaks differentiability. By bypassing sampling, the model can be trained directly on multi-step trajectories, enabling consistent autoregression during both training and inference.

  \item \textbf{Sequence Modeling (Transformer Backbone).}
        The probability distribution of each time series value, $\boldsymbol{p_i}$, is mapped to a corresponding embedding vector \(\boldsymbol{e}_i\). The stream \((\boldsymbol{e}_{1},\dots,\boldsymbol{e}_{t})\) is processed by a causal
        Transformer \(f_{\theta}\) that outputs logits
        \(\boldsymbol{z}_{t}=f_{\theta}(\boldsymbol{e}_{\le t})\in\mathbb{R}^{V}\),
        corresponding to the next-token distribution
        \(\boldsymbol{\hat{p}}_{t}=\text{softmax}(\boldsymbol{z}_{t})\).

  \item \textbf{Risk-Aware Decoding.}
        Assuming that the model is well-calibrated and makes accurate probabilistic predictions, the risk-aware decoding module calculates the point forecast that minimizes the expected task-specific risk, $\mathbb{E}[\mathcal{R}]$.
\end{enumerate}
Together, these components form a fully differentiable, autoregressive architecture that can be trained end-to-end on full forecast trajectories. This mitigates issues such as exposure bias and compounding errors typically introduced by teacher forcing. Moreover, \textsc{SoTra} aligns the model's outputs with downstream utility through its zone-aware decoding mechanism.

\subsection{Soft-Token Embedding}

Each token \(v \in \{1, \dots, V\}\) is associated with a learnable embedding vector \(E_v \in \mathbb{R}^d\), where \(E \in \mathbb{R}^{V \times d}\) denotes the embedding matrix. Given a predicted distribution \(\hat{\boldsymbol{p}}_t \in \Delta^{V-1}\), the soft embedding \(\boldsymbol{e}_t \in \mathbb{R}^d\) is computed as a weighted average:
\begin{equation}
    \boldsymbol{e}_t = E^\top \hat{\boldsymbol{p}}_t
\end{equation}
This formulation avoids sampling and preserves differentiability across the sequence. As a result, the model remains fully differentiable and supports training across full trajectories.

To enable soft-token embedding, we discretize the continuous time series into a vocabulary of \(V\) levels. Before discretization, the input time series is normalized using \emph{reversible instance normalization} \citep{kim2021reversible}, which normalizes each input segment based on its mean and standard deviation. This reduces distributional shifts between the training and test sets. After prediction, the transformation is inverted using the stored normalization statistics.

We then clamp the normalized values to \([-3\sigma, +3\sigma]\), assign any outliers to overflow bins, and divide the range into \(V\) uniform levels, each mapped to a token. This creates a compact representation for the model. To retain information about the original scale, we also tokenize the normalization mean and standard deviation and append their embeddings to the input, giving the model the needed context for accurate decoding.

\begin{figure}[t]
    \centering
    \includegraphics[width=1.00\linewidth]{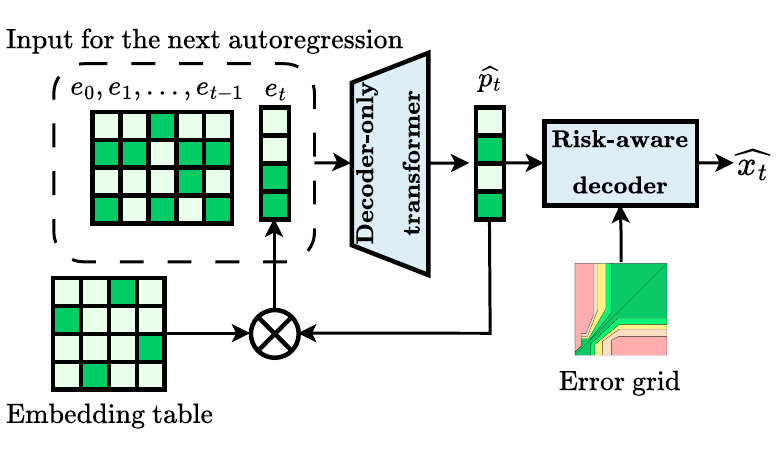}
    \caption{Overview of \textsc{SoTra}. At each autoregressive step, \textsc{SoTra} predicts the probability distribution over the next token, \(\boldsymbol{p}_t\), transforms this distribution into a soft embedding, and appends it to the sequence of past embeddings. This updated sequence is then used to predict the next distribution, \(\boldsymbol{p}_{t+1}\). \textsc{SoTra} produces continuous categorical forecasts by mapping predicted probability distributions directly to embeddings via matrix multiplication, eliminating the need for sampling and enabling fully differentiable trajectory generation. Final point forecasts are decoded by minimizing expected risk under an application-specific error grid, rather than using standard distance-based metrics.
}
    \label{fig:SoTra_overview}
\end{figure}
\subsection{Transformer Backbone}

\subsubsection{\textbf{Model.}} The sequence of soft embeddings \((\boldsymbol{e}_1, \dots, \boldsymbol{e}_t)\) is processed by a GPT-style decoder-only Transformer. The model width is set to match the vocabulary size, $V$. This ensures that the output logits \(\boldsymbol{z}_t \in \mathbb{R}^{V}\) can fully represent a probability distribution over all discretized levels without information loss.

\subsubsection{\textbf{Training.}}
We use a two-stage curriculum:
\begin{enumerate}
\item \emph{Stage 1 — Next-token pre-training.}  
      Using teacher forcing, the model is optimized in the standard next-token setting.  
      Each mini-batch provides \(T \times B\) parallel supervision signals, where \(B\) denotes the batch size.

\item \emph{Stage 2 — Trajectory fine-tuning.}  
      The network is subsequently unrolled for \(L\) future steps without teacher forcing.  
      At every step, the predicted distribution \(\hat{\boldsymbol{p}}_t\) is routed back through the soft-embedding layer, keeping the computation graph fully differentiable.  
      In this phase, each mini-batch performs \(L\) autoregressive passes and yields \(L \times B\) supervision signals.
\end{enumerate}
Because Stage 1 is considerably more efficient, we first pre-train the model on the next-token objective, then fine-tune it under the trajectory objective.

\subsection{Risk-Aware Decoding}

To produce a point estimate from the predicted token distribution \(\hat{\boldsymbol{p}}_t \in \Delta^{V-1}\), we use a decoding scheme that minimizes a combination of the expected application-specific zone-based risk and a regularizing expected MSE term.
Let \(\phi(v)\) denote the center of bin \(v\). The final decision rule is:

\begin{align}
\hat{x}_t = \phi \bigl( \arg\min_{x \in \{1, \dots, V\}} 
\sum_{v=1}^{V} \boldsymbol{\hat{p}}_{t,v} \big[
    & \lambda \cdot f_{r}(\phi(x), \phi(v)) \nonumber \\
    & + (\phi(x) - \phi(v))^2
\big] \bigl)
\label{eq:decode}
\end{align}
where \(\lambda\) is a tunable hyperparameter that balances the risk-aware objective with mean-squared error. Note that we make no assumptions about the shape or granularity of the zones; they may be arbitrarily defined and made sufficiently fine-grained to approximate continuous error grids.

\begin{table*}[t!]
  \centering
  \small
  \setlength{\tabcolsep}{2.75pt}
  \begin{tabular}{>{\centering\arraybackslash}m{0.5cm}| l  *{5}{|ccc}}
    \toprule
    \multirow{2}{*}{\rotatebox{90}{Dataset}} &
    \multirow{2}{*}{$L$} &
      \multicolumn{3}{c}{\textbf{SoTra}} &
      \multicolumn{3}{c}{\textbf{PatchTST}} &
      \multicolumn{3}{c}{\textbf{iTrans}} &
      \multicolumn{3}{c}{\textbf{Chronos}} &
      \multicolumn{3}{c}{\textbf{DLinear}}\\
    \cmidrule(lr){3-5}\cmidrule(lr){6-8}\cmidrule(lr){9-11}\cmidrule(lr){12-14}\cmidrule(lr){15-17}
      & & Risk & Risky\% & RMSE & Risk & Risky\% & RMSE & Risk & Risky\% & RMSE & Risk & Risky\% & RMSE & Risk & Risky\% & RMSE\\
    \midrule
\multirow{5}{*}{\rotatebox{90}{\textbf{DCLP3}}} 
& 6   & \textbf{0.112} & \textbf{0.246} & 16.90 & 0.325 & 1.400 & 17.75 & \underline{0.153} & \underline{0.460} & \underline{16.75} & 0.185 & 0.570 & 17.81 & 0.181 & 0.650 & \textbf{16.49}\\
& 12  & \textbf{0.341} & \textbf{1.003} & 27.05 & 0.550 & 2.170 & \underline{26.27} & 0.461 & 1.750 & \textbf{25.76} & 0.488 & 1.710 & 29.05 & \underline{0.455} & \underline{1.640} & 26.38\\
& 24  & \textbf{0.735} & \textbf{2.389} & 38.57 & 0.906 & 3.610 & \textbf{36.71} & \underline{0.863} & \underline{3.410} & \underline{37.06} & 1.016 & 4.100 & 43.05 & 0.886 & 3.440 & 37.94\\
& 48  & \textbf{1.144} & \textbf{3.538} & 49.00 & \underline{1.345} & \underline{5.710} & \textbf{45.37} & 1.347 & 5.730 & \underline{45.95} & 1.611 & 7.150 & 55.33 & 1.386 & 5.810 & 46.58\\
& Avg & \textbf{0.583} & \textbf{1.794} & 32.88 & 0.782 & 3.222 & \underline{31.52} & \underline{0.706} & \underline{2.837} & \textbf{31.13} & 0.824 & 3.382 & 36.31 & 0.727 & 2.885 & 31.84\\
    \midrule
\multirow{5}{*}{\rotatebox{90}{\textbf{PEDAP}}} 
& 6   & \textbf{0.201} & \textbf{0.613} & 19.87 & 0.560 & 2.560 & 19.73 & 0.288 & 1.130 & \textbf{19.03} & \underline{0.268} & \underline{0.920} & 19.98 & 0.325 & 1.290 & \underline{19.24}\\
& 12  & \textbf{0.526} & \textbf{1.697} & 32.08 & 0.789 & 3.330 & \textbf{29.27} & 0.697 & 2.810 & \underline{29.42} & \underline{0.696} & \underline{2.640} & 32.87 & 0.701 & 2.760 & 30.20\\
& 24  & \textbf{1.019} & \textbf{3.537} & 44.14 & 1.259 & 5.460 & \textbf{40.41} & 1.218 & 5.250 & \underline{41.02} & 1.339 & 5.710 & 48.05 & \underline{1.218} & \underline{5.100} & 41.97\\
& 48  & \textbf{1.415} & \textbf{4.448} & 53.20 & 1.747 & 7.800 & \underline{48.59} & \underline{1.714} & 7.670 & \textbf{48.49} & 2.010 & 9.190 & 59.53 & 1.726 & \underline{7.570} & 50.03\\
& Avg & \textbf{0.791} & \textbf{2.574} & 37.32 & 1.089 & 4.787 & \underline{34.50} & \underline{0.979} & 4.215 & \textbf{34.49} & 1.078 & 4.615 & 40.10 & 0.992 & \underline{4.180} & 35.36\\
    \midrule
\multirow{5}{*}{\rotatebox{90}{\textbf{MBP}}} 
& 6   & \textbf{1.047} & \textbf{0.382} & \textbf{5.587} & 1.062 & 0.536 & \underline{5.845} & \underline{1.054} & \underline{0.535} & 5.868 & 1.064 & 0.633 & 6.332 & 1.062 & 0.570 & 6.095\\
& 12  & \textbf{1.064} & \textbf{0.486} & \textbf{6.522} & 1.087 & 0.624 & \underline{6.723} & \underline{1.079} & \underline{0.621} & 6.772 & 1.097 & 0.818 & 7.645 & 1.093 & 0.699 & 7.063\\
& 24  & \textbf{1.103} & \textbf{0.732} & \underline{8.120} & 1.130 & \underline{0.822} & \textbf{8.048} & \underline{1.123} & 0.870 & 8.155 & 1.155 & 1.199 & 9.304 & 1.142 & 0.955 & 8.413\\
 & 48  & 1.196 & 1.449 & 10.61 & \underline{1.192} & \textbf{1.225} & \underline{9.475} & \textbf{1.191} & \underline{1.251} & \textbf{9.559} & 1.241 & 2.060 & 11.438 & 1.212 & 1.378 & 9.757\\
 & Avg & \textbf{1.103} & \textbf{0.762} & 7.710 & 1.118 & \underline{0.802} & \textbf{7.522} & \underline{1.112} & 0.819 & \underline{7.558} & 1.139 & 1.177 & 8.679 & 1.127 & 0.900 & 7.831\\
    \midrule
\multirow{5}{*}{\rotatebox{90}{\textbf{SBP}}} 
& 6   & \textbf{1.038} & \textbf{0.318} & \textbf{6.821} & \underline{1.044} & \underline{0.437} & \underline{7.165} & 1.047 & 0.469 & 7.254 & 1.053 & 0.580 & 8.007 & 1.047 & 0.516 & 7.466\\
& 12  & \textbf{1.051} & \textbf{0.445} & \underline{8.486} & \underline{1.065} & \underline{0.620} & \textbf{8.429} & 1.067 & 0.650 & 8.546 & 1.079 & 0.864 & 9.698 & 1.069 & 0.711 & 8.921\\
& 24  & \textbf{1.075} & \textbf{0.740} & 10.79 & 1.099 & \underline{1.000} & \textbf{10.459} & \underline{1.099} & 1.024 & \underline{10.566} & 1.130 & 1.553 & 12.469 & 1.114 & 1.183 & 10.925\\
& 48  & \textbf{1.116} & \textbf{1.247} & 13.66 & 1.164 & 1.898 & \textbf{12.661} & \underline{1.152} & 1.770 & \underline{12.776} & 1.208 & 2.887 & 15.653 & 1.162 & \underline{1.880} & 13.047\\
& Avg & \textbf{1.070} & \textbf{0.688} & 9.940 & 1.093 & 0.989 & \textbf{9.678} & \underline{1.091} & \underline{0.979} & \underline{9.785} & 1.117 & 1.471 & 11.456 & 1.098 & 1.072 & 10.089\\
    \midrule
\multicolumn{2}{c|}{\textbf{Win count}} &
  \textbf{19} & \textbf{19} & 3 & 0 & 1 & \textbf{10} & 1 & 0 & 6 & 0 & 0 & 0 & 0 & 0 & 1\\
  \midrule
  \end{tabular}
  \caption{Multi-horizon forecasting results. \textbf{Bold} indicates the best performance, and \underline{underline} indicates the second-best performance.}
  \label{tab:results_metrics}
\end{table*}

\section{Experiments}
We comprehensively evaluate \textsc{SoTra} in safety-critical clinical forecasting settings. Our results demonstrate that \textsc{SoTra} achieves state-of-the-art performance in terms of clinical risk, while remaining competitive with existing methods in RMSE. Furthermore, we evaluate the generalizability of \textsc{SoTra} by testing it on an unseen dataset within the same clinical domain.

\subsubsection{\textbf{Datasets and Preprocessing.}} We evaluate our method on five clinical time series datasets. For glucose forecasting, we use DCLP3 \citep{brown2019six}, and PEDAP \citep{wadwa2023trial}, each comprising continuous glucose monitoring (CGM) traces collected from individuals with type 1 diabetes at five-minute sampling intervals. For blood pressure forecasting, we rely on the VitalDB dataset \citep{lee2022vitaldb}, which contains high-resolution arterial and non-invasive blood pressure recordings from surgical patients. We extract the systolic blood pressure (SBP) and mean blood pressure (MBP). Following prior work \citep{baek2023cuffless}, we exclude measurements with physiologically implausible values and downsample all series to 0.1 Hz, facilitating longer-term prediction while controlling the effective sequence length.

\subsubsection{\textbf{Forecasting Tasks.}} We train models to forecast future values at horizons of 6, 12, 24, and 48 time steps, similar to \citet{karagoz2025comparative}. The forecasting horizon corresponds to 0.5, 1, 2, and 4 hours for the glucose datasets, and 1, 2, 4, and 8 minutes for the blood pressure datasets. Each forecast is conditioned on a fixed-length history: 288 samples (24 hours) for glucose, and 180 samples (30 minutes) for blood pressure.

Our primary evaluation metric is \textit{clinical risk}. For glucose forecasting, we use the Clarke Error Grid \citep{clarke1987evaluating}. The numerical risk values assigned to Clarke Error Grid zones can vary substantially depending on the clinical context and application. This is reflected in the evolution from Clarke's original qualitative zones to the Surveillance Error Grid's continuous scoring scheme \citep{klonoff2014surveillance}, as well as implementation-specific values such as those used in the CRAN \texttt{ega} package \citep{ega}. These variations highlight that risk assessment must take into account the clinical setting, patient population, and the relative consequences of hypoglycemic versus hyperglycemic errors. In our work, following consultation with diabetes experts at our institution, we assign risk scores of 0, 1, 7.5, 17.5, and 37.5 to zones A, B, C, D, and E, respectively. For blood pressure, we employ the arterial pressure error grid proposed by \citet{saugel2018error}, using the risk stratification annotations proposed by \citet{juri2021error}. We also report the percentage of risky forecasts—those falling outside zones A and B of the error grid—as well as RMSE.

\subsubsection{\textbf{Baselines.}} We compare our method against a set of strong baselines spanning simple linear, transformer-based, and foundation models. \textbf{DLinear} is an embarrassingly simple linear model that has been shown to outperform more complex architectures on a range of forecasting benchmarks \citep{zeng2023transformers}. \textbf{iTransformer} \citep{liu2023itransformer} is a transformer-based model that has demonstrated competitive performance on blood glucose prediction tasks \citep{karagoz2025comparative}. \textbf{PatchTST} \citep{Yuqietal-2023-PatchTST} represents the state-of-the-art in transformer-based forecasting and uses a patching mechanism to enhance long-term temporal modeling. Finally, \textbf{Chronos} \citep{ansari2024chronos} is a foundation model trained on large-scale time series data that supports zero-shot probabilistic forecasting. We used the base version of Chronos.
%––––– Ablation table –––––
\begin{table*}[t!]
  \centering
  \setlength{\tabcolsep}{3pt}
  \small
  \begin{tabular}{c | c | c *{4}{|ccc}}
    \toprule
    \multirow{2}{*}{Config} &
      \multirow{2}{*}{\begin{tabular}{@{}c@{}}Soft token\\trajectory\end{tabular}} &
      \multirow{2}{*}{Risk-aware} &
      \multicolumn{3}{c}{\textbf{DCLP3}} &
      \multicolumn{3}{c}{\textbf{PEDAP}} &
      \multicolumn{3}{c}{\textbf{MBP}} &
      \multicolumn{3}{c}{\textbf{SBP}}\\
    \cmidrule(lr){4-6}\cmidrule(lr){7-9}\cmidrule(lr){10-12}\cmidrule(lr){13-15}
      & & & Risk & Risky\% & RMSE & Risk & Risky\% & RMSE & Risk & Risky\% & RMSE & Risk & Risky\% & RMSE\\
    \midrule
    \textcircled{1} & \cmark & \cmark & \textbf{0.583}& \textbf{1.794}& \underline{32.88}& \textbf{0.791} & \textbf{2.574}  & \underline{37.32} & \textbf{1.103} & \textbf{0.762} & \underline{7.710}  & \textbf{1.070} & \textbf{0.688} & \underline{9.940} \\
    \textcircled{2} & \cmark & \xmark & \underline{0.768} &\underline{3.204}&    \textbf{31.51}& \underline{1.057} & \underline{4.627}  & \textbf{35.75} & \underline{1.116} & \underline{0.823} & \textbf{7.488}  & \underline{1.089} & \underline{0.920} & \textbf{9.542} \\
    \textcircled{3} & \xmark & \cmark &         1.416 &         7.255 &             49.88& 2.364 & 13.992 & 67.52 & 1.121 & 0.834 & 8.125 & 1.098 & 1.036 & 10.690\\
    \textcircled{4} & \xmark & \xmark &         1.368 &         7.422 &             49.54& 2.285 & 14.329 & 67.33 & 1.129 & 0.880 & 8.013 & 1.107 & 1.208 & 10.548\\
    \bottomrule
  \end{tabular}
  \caption{Ablation study on SoTra.  Four configurations vary in the use of soft-token decoding and risk-aware training.}
  \label{tab:ablation}
\end{table*}
\subsection{Main Results}
Table~\ref{tab:results_metrics} summarizes forecasting performance. Overall, \textsc{SoTra} substantially reduces clinical risk: across 20 comparisons, it achieves the lowest risk and the lowest percentage of risky predictions in 19 cases. On the glucose datasets, it delivers an average of 18\% lower clinical risk and 32\% fewer risky forecasts than the best baselines, iTransformer and PatchTST, respectively.

For blood pressure forecasting, \textsc{SoTra} achieves 1.3\% lower clinical risk than iTransformer and 24\% fewer risky forecasts than PatchTST. Because the risk scheme of \citet{juri2021error} includes a baseline penalty of one even for perfect predictions, subtracting this offset reveals that SoTra reduces the clinical risk by approximately 15\% compared to the best baseline.

Although \textsc{SoTra} does not obtain the best RMSE, it remains competitive—on average, only 4.7\% worse than the strongest baseline across datasets and horizons. Given that it is trained with cross-entropy and optimized for clinical risk rather than RMSE, this represents a favorable trade-off for deployment in safety-critical applications.

\begin{figure}[t!]
  \centering
  %–––– Row 1 ––––
  \begin{subfigure}[t]{0.48\linewidth}
    \centering
    \includegraphics[width=\linewidth]{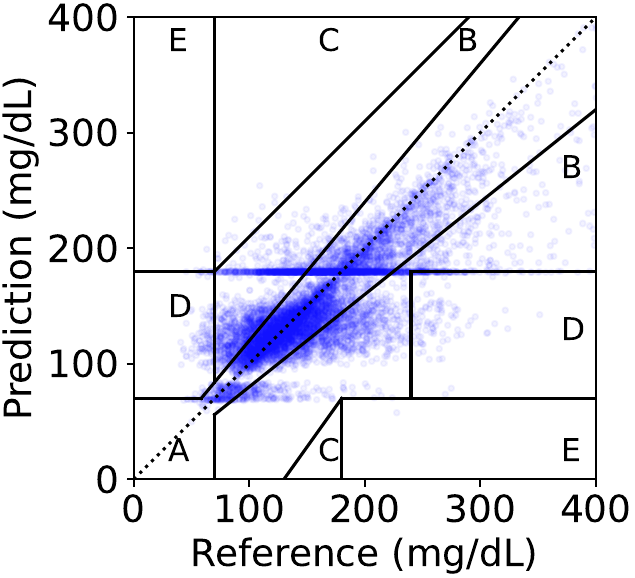}
    \caption{DCLP3 dataset, RA on.}   % (a)
    \label{fig:dclp3_ra}
  \end{subfigure}\hfill
  \begin{subfigure}[t]{0.48\linewidth}
    \centering
    \includegraphics[width=\linewidth]{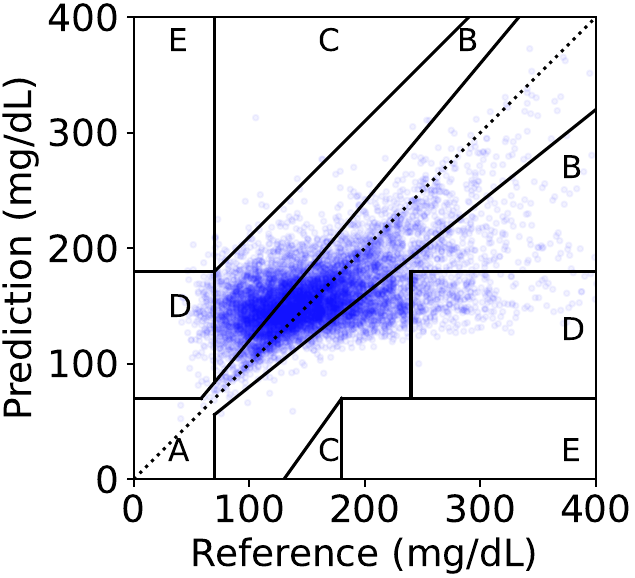}
    \caption{DCLP3 dataset, RA off.}  % (b)
    \label{fig:dclp3_nra}
  \end{subfigure}

  \vspace{0.5em}

  %–––– Row 2 ––––
  \begin{subfigure}[t]{0.48\linewidth}
    \centering
    \includegraphics[width=\linewidth]{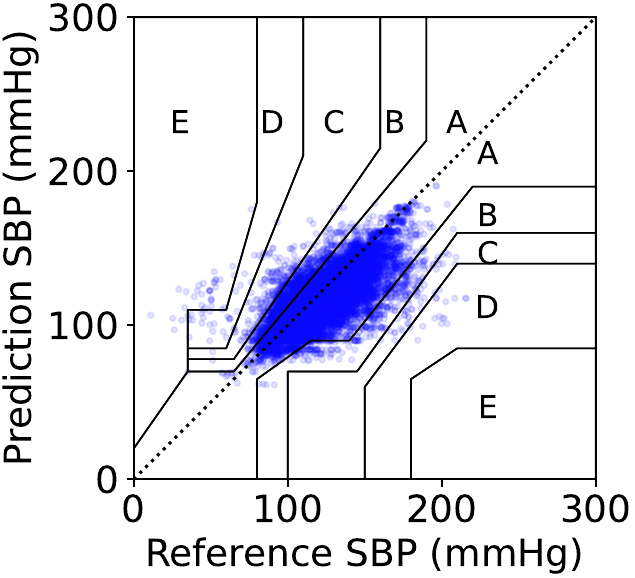}
    \caption{SBP dataset, RA on.}     % (c)
    \label{fig:sbp_ra}
  \end{subfigure}\hfill
  \begin{subfigure}[t]{0.48\linewidth}
    \centering
    \includegraphics[width=\linewidth]{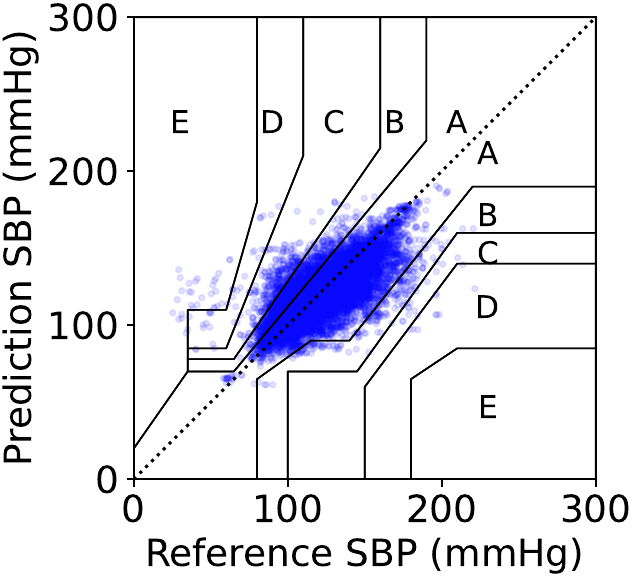}
    \caption{SBP dataset, RA off.}    % (d)
    \label{fig:sbp_nra}
  \end{subfigure}

  \caption{Error-grid impact of risk-aware decoding (RA) for a prediction horizon of $L=48$. Enabling RA visibly reduces the density of points in clinically higher-risk zones (e.g., Zone D in DCLP3, Zone C in SBP).}
  \label{fig:error_grids}
\end{figure}

\begin{table}[t!]
  \centering
  \begin{tabular}{@{}l|cc|ccc@{}}
    \toprule
        Dataset & \multicolumn{2}{c}{\textbf{DCLP3}} &
          \multicolumn{2}{c}{\textbf{SBP}} \\ \cmidrule(lr){2-3}\cmidrule(lr){4-5}
       \textsc{SoTra config} & \textcircled{1} & \textcircled{2} & \textcircled{1} & \textcircled{2} \\ \midrule
    \textbf{Zone A (\%)} & 54.10 & 55.28 & 93.22 & 90.52 \\ 
    \midrule
    \textbf{Zone B (\%)} & 42.35 & 38.61 & 5.52  & 7.76  \\ 
    \midrule
    \textbf{Zone C (\%)} & 0.06  & 0.30  & 0.95  & 1.41  \\ 
    \midrule
    \textbf{Zone D (\%)} & 2.92  & 5.68  & 0.17  & 0.16  \\
    \midrule
    \textbf{Zone E (\%)} & 0.54  & 0.11  & 0.11  & 0.13  \\
    \bottomrule
  \end{tabular}
  \caption{Percentage of predictions falling into each error grid zone for two \textsc{SoTra} configurations.}
  \label{tab:ra_ablatio}
\end{table}

\subsection{Ablation}
We ablate the two core ingredients—\textit{soft-token trajectory training} and \textit{risk-aware decoding}—by toggling each on or off.

\textbf{Risk-aware off:} We set $\lambda{=}0$ in Eq.~\ref{eq:decode}, reducing decoding to expected MSE minimization.

\textbf{Soft token trajectory training off:} We use the model to predict the next token. Following \citep{ansari2024chronos, liu2025sundial}, we sample 5 values from the predicted distribution and use the median as the next autoregressive input. Decoding is based on either MSE or risk minimization, depending on whether risk awareness is disabled or enabled.

We first examine the clinical impact of risk-aware decoding in \textsc{SoTra}. Fig.~\ref{fig:error_grids} shows Clarke error grids with and without risk-aware decoding for a prediction horizon of 48, and Table~\ref{tab:ra_ablatio} reports the percentage of predictions falling into each zone under the same setting. On the DCLP3 dataset, disabling risk-aware decoding increases the proportion of predictions in Zone D from 2.92\% to 5.68\%, indicating a higher incidence of failures to detect potentially dangerous hypo- or hyperglycemic events \citep{clarke1987evaluating}. On the SBP dataset, the percentage of predictions in Zone C rises from 0.95\% to 1.41\%, reflecting an increased likelihood of errors that could lead to unnecessary interventions with moderate, though not life-threatening, clinical consequences \citep{juri2021error}.

Fig.~\ref{fig:firstpage} illustrates two predicted trajectories for the same example: forecast A, generated with risk-aware decoding enabled, and forecast B, with risk-awareness disabled. While forecast A yields a worse RMSE, it correctly identifies a hypoglycemic episode, demonstrating the value of optimizing for clinical risk rather than statistical accuracy alone.

Table~\ref{tab:ablation} reports results averaged over horizons 6, 12, 24, and 48. Trajectory training enabled by soft tokens consistently improves RMSE across all datasets and reduces risk in all datasets, highlighting its effectiveness in mitigating teacher forcing. When trajectory training is enabled, risk-aware decoding further reduces risk in every case. However, this effect disappears without trajectory training, as risk-aware decoding depends on the accuracy of the model's predicted probability distribution. The best RMSE is achieved with trajectory training enabled and risk-aware decoding disabled—just 0.7\% worse than the strongest baseline.

\subsection{Probabilistic Forecasting}

We evaluate the quality of the predictive distributions using the Continuous Ranked Probability Score (CRPS) and calibration curves.  
Our comparison focuses on three probabilistic forecasters:  
(i) \textsc{SoTra} with trajectory training (full model),  
(ii) \textsc{SoTra} without trajectory training, and  
(iii) Chronos, the only probabilistic baseline.

Table~\ref{tab:crps_results} reports CRPS averaged over all horizons.  
Across all datasets, \textsc{SoTra} with trajectory training achieves the best probabilistic accuracy. Removing trajectory training consistently worsens CRPS, showing that exposure-bias mitigation also improves distributional quality. \textsc{SoTra} also outperforms Chronos, a strong probabilistic baseline.

\begin{table}[t]
\centering
\small
\begin{tabular}{lccc}
\toprule
\textbf{Dataset} & \textbf{SoTra (traj.)} & \textbf{SoTra (no-traj.)} & \textbf{Chronos} \\
\midrule
DCLP3      & 15.83 & 29.54 & 17.08 \\
PEDAP      & 17.89 & 41.47 & 19.28 \\
MBP        & 4.51 & 5.60 & 5.00 \\
SBP        & 3.41 & 3.97 & 3.93 \\
\bottomrule
\end{tabular}
\caption{Average CRPS across forecasting horizons.}
\label{tab:crps_results}
\end{table}

\begin{figure}[t!]
  \centering
  %–––– Row 1 ––––
  \begin{subfigure}[t]{0.48\linewidth}
    \centering
    \includegraphics[width=\linewidth]{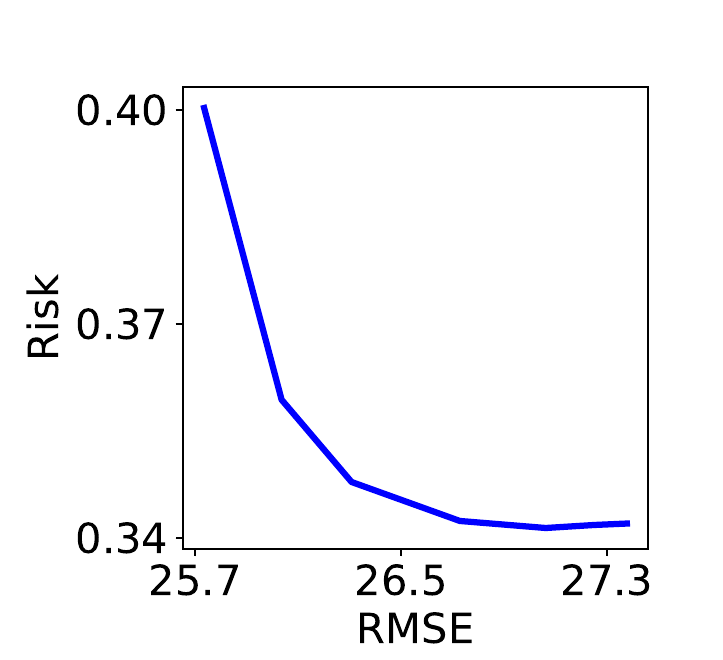}
    \caption{DCLP3 dataset.}   % (a)
  \end{subfigure}\hfill
  \begin{subfigure}[t]{0.48\linewidth}
    \centering
    \includegraphics[width=\linewidth]{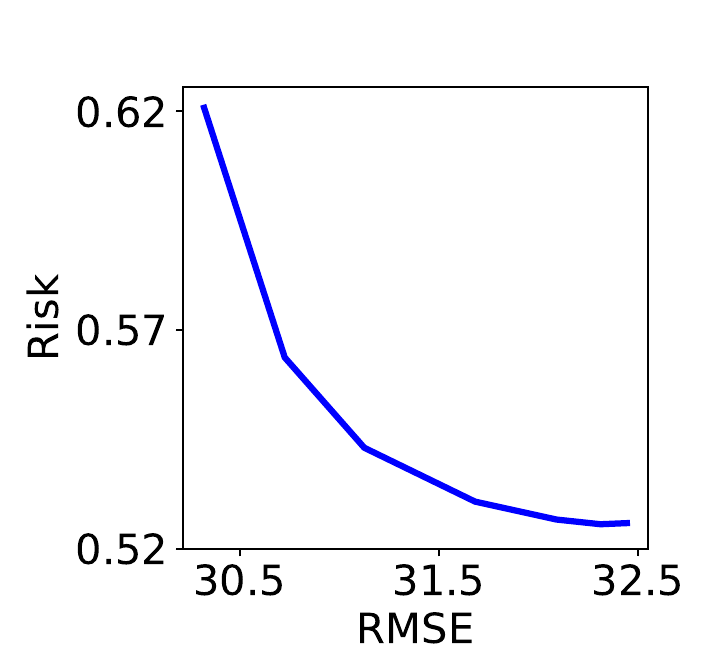}
    \caption{PEDAP dataset.}  % (b)
  \end{subfigure}

  \vspace{0.5em}

  %–––– Row 2 ––––
  \begin{subfigure}[t]{0.48\linewidth}
    \centering
    \includegraphics[width=\linewidth]{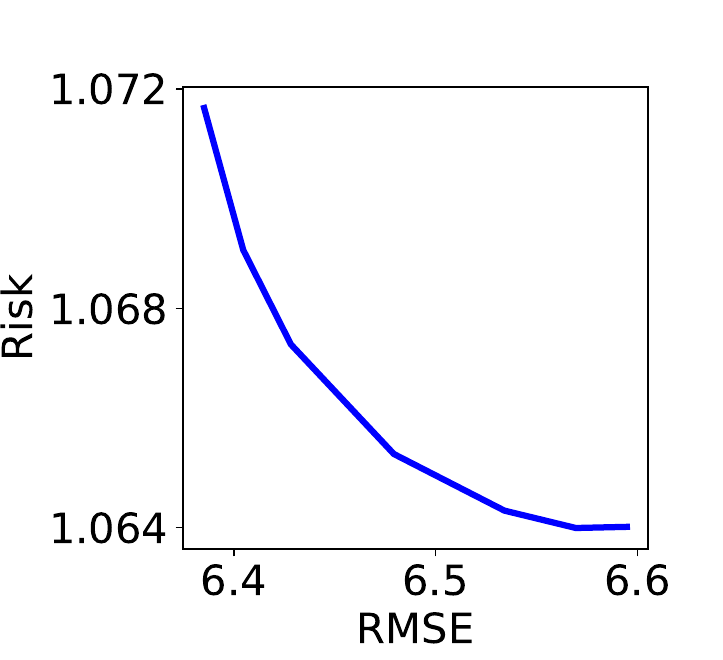}
    \caption{MBP dataset.}     % (c)
  \end{subfigure}\hfill
  \begin{subfigure}[t]{0.48\linewidth}
    \centering
    \includegraphics[width=\linewidth]{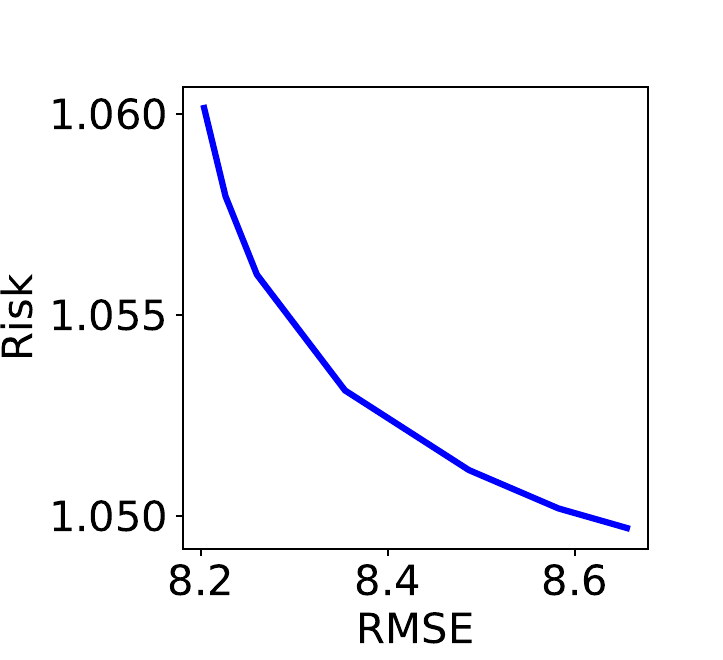}
    \caption{SBP dataset.}    % (d)
  \end{subfigure}

  \caption{Effect of the hyperparameter $\lambda$.}
  \label{fig:lambda_sweep}
\end{figure}

\subsection{Hyperparameters}

Model hyperparameters were selected on the DCLP3 dataset by minimizing RMSE with risk-aware decoding disabled, separating representational capacity from task-specific risk weighting. The final architecture uses a model width of 256, four layers, and four attention heads; this configuration was used across all datasets without further tuning.

Training follows a two-stage procedure: (i) teacher-forced next-token training with a learning rate of $10^{-4}$, and (ii) soft-token trajectory training with a reduced learning rate of $10^{-5}$ due to noisier gradient estimates. Early stopping and gradient clipping (norm 1.0) were applied for stability.

To study the effect of the clinical risk coefficient $\lambda$ in Eq.~\ref{eq:decode}, we swept $\lambda$ from 3 to 200 and plotted RMSE against clinical risk (Fig.~\ref{fig:lambda_sweep}). The curves show a clear trade-off: increasing $\lambda$ reduces clinical risk while gradually increasing RMSE.

All experiments were implemented in PyTorch and run with mixed precision on NVIDIA A100/A6000-class GPUs.

\section{Discussion}

\subsubsection{\textbf{Soft-token trajectory training.}}
Soft-token propagation enables differentiable trajectory-level training, substantially reducing exposure bias in autoregressive forecasting. This improves multi-step stability—a key requirement for predictive control—and allows the model to anticipate and correct error accumulation across the horizon.

\subsubsection{\textbf{Risk-aware decoding.}}
By separating probability modeling from risk-based decision-making, SoTra provides a principled way to optimize the forecast that is actually used by downstream controllers. Minimizing expected clinical risk yields markedly safer predictions (e.g., 18\% and 14\% risk reductions in glucose and blood pressure), while maintaining competitive RMSE. This allows forecasts to align with zone-based safety criteria already used in clinical decision support and emerging closed-loop systems.

\subsubsection{\textbf{Implications for control.}}
The ability to produce calibrated, risk-aware multi-step predictions makes SoTra directly relevant for MPC-based regulation of physiological variables, such as glucose or blood pressure. Because MPC optimizes over predicted trajectories, improved distributional calibration and zone-aware objectives can translate into more conservative and stable control actions near dangerous states. The modularity of the soft-token approach also enables integration with multivariate models or foundation-model encoders, offering a path toward safe and adaptive closed-loop controllers.

\section{Conclusion}
Soft-Token Trajectory Forecasting (\textsc{SoTra}) addresses core challenges in safety-critical time series prediction by mitigating exposure bias and enabling risk-aware forecasting. Instead of sampling discrete tokens, \textsc{SoTra} maintains and propagates continuous uncertainty distributions, allowing forecasts to align with clinically meaningful objectives. 

Evaluations on glucose and blood pressure prediction tasks show that \textsc{SoTra} reliably reduces high-risk forecast errors at a small cost to the traditional accuracy metrics. These findings underscore the importance of modeling uncertainty explicitly in autoregressive settings, particularly when decision quality depends on asymmetric risk.

To facilitate future work in this direction, we will release our code and zone-based evaluation protocols for risk-sensitive time series forecasting.

\section*{DECLARATION OF GENERATIVE AI AND AI-ASSISTED TECHNOLOGIES IN THE WRITING PROCESS}
During the preparation of this work, the authors used ChatGPT for assistance with language editing, restructuring, and improving the clarity of the manuscript text. All technical content, experimental results, and scientific conclusions were developed by the authors. After using this tool, the authors reviewed and edited the content as needed and take full responsibility for the final version of the publication.

\bibliography{ifacconf}             % bib file to produce the bibliography
                                                     % with bibtex (preferred)
                                                   
%\begin{thebibliography}{xx}  % you can also add the bibliography by hand

%\bibitem[Able(1956)]{Abl:56}
%B.C. Able.
%\newblock Nucleic acid content of microscope.
%\newblock \emph{Nature}, 135:\penalty0 7--9, 1956.

%\bibitem[Able et~al.(1954)Able, Tagg, and Rush]{AbTaRu:54}
%B.C. Able, R.A. Tagg, and M.~Rush.
%\newblock Enzyme-catalyzed cellular transanimations.
%\newblock In A.F. Round, editor, \emph{Advances in Enzymology}, volume~2, pages
%  125--247. Academic Press, New York, 3rd edition, 1954.

%\bibitem[Keohane(1958)]{Keo:58}
%\newblock \emph{Power and Interdependence: World Politics in Transitions}.
%\newblock Little, Brown \& Co., Boston, 1958.

%\bibitem[Powers(1985)]{Pow:85}
%\newblock Is there a way out?
%\newblock \emph{Harpers}, pages 35--47, June 1985.

%\bibitem[Soukhanov(1992)]{Heritage:92}
%\newblock \emph{{The American Heritage. Dictionary of the American Language}}.
%\newblock Houghton Mifflin Company, 1992.

%\end{thebibliography}
\end{document}